\pdfoutput=1

\documentclass[sigconf, authorversion]{acmart}
\usepackage{listings}
\usepackage{nameref}

\AtBeginDocument{%
  \providecommand\BibTeX{{%
    \normalfont B\kern-0.5em{\scshape i\kern-0.25em b}\kern-0.8em\TeX}}}

\copyrightyear{2020}
\acmYear{2020}
\setcopyright{rightsretained}
\acmConference[SIGSPATIAL '20]{28th International Conference on Advances in Geographic Information Systems}{November 3--6, 2020}{Seattle, WA, USA}
\acmBooktitle{28th International Conference on Advances in Geographic Information Systems (SIGSPATIAL '20), November 3--6, 2020, Seattle, WA, USA}
\acmDOI{10.1145/3397536.3422252}
\acmISBN{978-1-4503-8019-5/20/11}

\acmPrice{15.00}

\begin{document}

\title{Predictive Collision Management for Time and Risk Dependent Path Planning}

\author{Carsten Hahn}
\affiliation{\institution{LMU Munich}}
\email{carsten.hahn@ifi.lmu.de}

\author{Sebastian Feld}
\affiliation{\institution{LMU Munich}}
\email{sebastian.feld@ifi.lmu.de}

\author{Hannes Schroter}
\affiliation{\institution{LMU Munich}}
\email{hannes@schroter.biz}

\renewcommand{\shortauthors}{Hahn et al.}

\begin{abstract}
Autonomous agents such as self-driving cars or parcel robots need to recognize and avoid possible collisions with obstacles in order to move successfully in their environment. Humans, however, have learned to predict movements intuitively and to avoid obstacles in a forward-looking way.
The task of collision avoidance can be divided into a global and a local level. Regarding the global level, we propose an approach called ``Predictive Collision Management Path Planning'' (PCMP). At the local level, solutions for collision avoidance are used that prevent an inevitable collision. Therefore, the aim of PCMP is to avoid unnecessary local collision scenarios using predictive collision management.
PCMP is a graph-based algorithm with a focus on the time dimension consisting of three parts: (1) movement prediction, (2) integration of movement prediction into a time-dependent graph, and (3) time and risk-dependent path planning.
The algorithm combines the search for a shortest path with the question: is the detour worth avoiding a possible collision scenario?
We evaluate the evasion behavior in different simulation scenarios and the results show that a risk-sensitive agent can avoid $47.3\%$ of the collision scenarios while making a detour of $1.3\%$. A risk-averse agent avoids up to $97.3\%$ of the collision scenarios with a detour of $39.1\%$. Thus, an agent's evasive behavior can be controlled actively and risk-dependent using PCMP. 
\vspace*{1cm}
\end{abstract}

\ccsdesc[500]{Computing methodologies~Planning under uncertainty}

\keywords{Time and Risk Dependent Path Planning,
	Trajectory Prediction,
	Artificial Neural Networks,
	Autonomous Agents}

\maketitle

\section{Introduction}
\label{sec:introduction}

Autonomous agents like self-driving cars or parcel robots are increasingly integrated into our everyday lives. A crucial step in their utilization is the ability to successfully navigate inside their environment. They need to find a path to their destination, but also have to recognize obstacles and avoid them.

Path planning, a subdomain of artificial intelligence, can typically be divided into several sub-problems \cite{Hahn2018}. The first part deals with making an environment with unlimited movement possibilities computable for a machine. Discretization methods are used that translate an agent's environment into a graph. If such a graph is available, path planning algorithms are used that calculate an optimal route between two points. 

Furthermore, the individual steps can be divided into a global and local level. At the global level, the agent is looking for a solution to the path planning problem, whereas static obstacles have already been taken into account in the discretization part. During the execution of the solution found, the agent may encounter dynamic obstacles. Preventing a collision is often called collision avoidance, and only the immediate environment and not the entire graph is relevant here. Evasion behavior is carried out at a local level with \emph{Potential Field} approaches \cite{Khatib1985} being common examples found in literature \cite{Montiel2015}. It may make sense for the agent to completely avoid the use of local procedures. If the implemented solution is inefficient or unreliable, it should only be used when absolutely necessary.

In this paper, we propose a novel approach for \emph{Predictive Collision Management Path Planning} (PCMP). It complements local strategies and enables them to be used in a sensible way. Obstacle's future movements are predicted using \emph{Long Short-Term Memory} (LSTM) models. The agent therefore knows which paths are likely to lead to a collision. This knowledge is uncertain information and thus poses a risk to the agent. The predictions' usefulness only arise when the agent is ready to take a certain risk of collision \cite{Philippsen2006}.

We integrate the predictions into a time-dependent graph and use a space-time extended A* algorithm to search for the optimal path, i.e. a compromise between risk of collision and path length. PCMP extends the agent's planning decision with the question: Is the detour worth it to avoid a local collision scenario? The answer depends on the agent's risk aversion and the cost of alternative paths. Our approach contains a risk parameter $r$, through which the agent's risk aversion and evasive behavior can be controlled.

\begin{figure}[htb]
    \centering
    \includegraphics[width=0.9\linewidth]{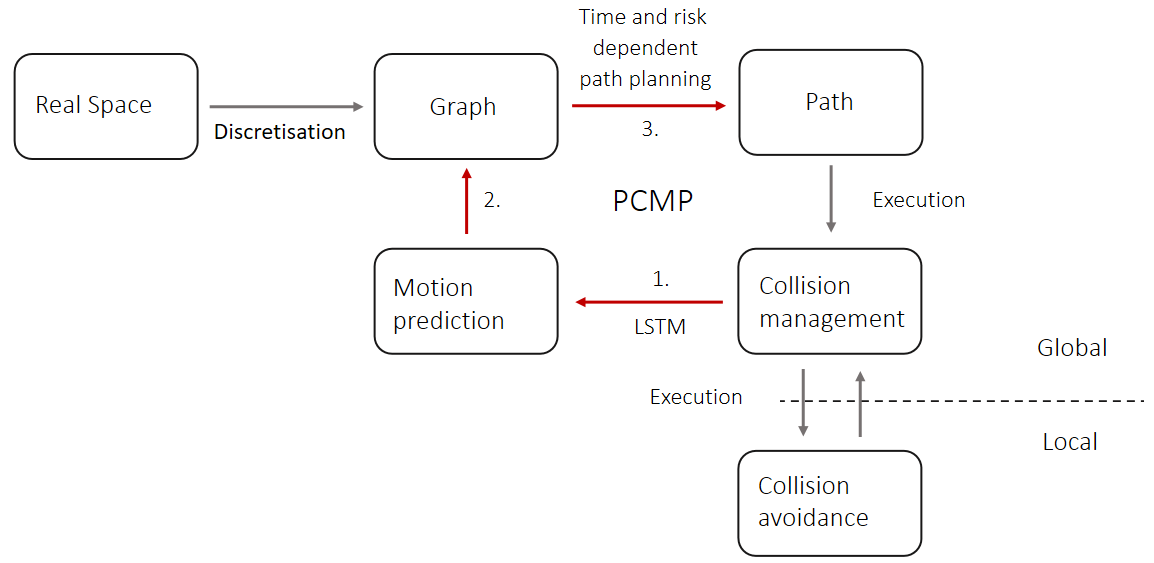}
    \caption{Placement of PCMP into path planning tasks.}
    \label{fig:pathplanningpcmp}
\end{figure}

Fig. \ref{fig:pathplanningpcmp} shows PCMP's different components and how the approach is to be placed in the domain of path planning. Arrows marked red form the generic framework of PCMP and also this paper's main structure: (1) motion predictions of dynamic obstacles using LSTM models, (2) integration of motion predictions in graphs, (3) time and risk-dependent path planning.

The paper is structured as follows: we highlight related work in Sec. \ref{sec:related} and elaborate PCMP's concept in Sec. \ref{sec:pcmp} with the underlying scenario (\ref{sec: scenario}), prediction (\ref{sec: prediction models}), graph integration (\ref{sec: integration}), path planning (\ref{sec: algo}), and a description of PCMP's collision control (\ref{subsec: risk parameter}). We evaluate our concept in Sec. \ref{sec: eval} and conclude in Sec. \ref{sec:conclusion}.

\section{Related Work}
\label{sec:related}

Collision avoidance is a broad field of research with a variety of different approaches and a speed with which new solutions are developed. We now provide a brief overview on prediction-based collision avoidance.

A central element is motion prediction of dynamic obstacles. Unlike \emph{Cooperative Pathfinding} problems \cite{Silver2005}, predictions are uncertain information. \cite{Philippsen2006} state that predictions are probability statements and therefore their benefits are only revealed if a certain collision risk is taken. \cite{Philippsen2006} also propose a multi-stage concept where collision avoidance algorithms such as \emph{Potential Field} \cite{Khatib1985} are used at a local level to ultimately prevent a collision.
Furthermore, a wide range of motion prediction techniques exist \cite{Lefevre2014}. The methods range from simple motion interpolations to complex, learning models. By using Kalman Filter, uncertainties in the predictions can be better estimated. They find use, among other concepts, in autonomous vehicles \cite{Ammoun2009,Barrios2011}. Bayesian filter and Monte Carlo methods are used to analyze movement behavior in road traffic \cite{Kaempchen2004,Eidehall2008}. The central task of motion prediction is to learn and predict future trajectories. In this context, Gaussian processes and Bayesian networks are important tools and describe integral parts of autonomous agents \cite{Gindele2015,Tran2014}. \cite{Kim2017} indicate that these models are characterized by a high level of complexity together with a high proportion of manual parameter settings. In a highly dynamic environment, this property may be a strong disadvantage. In contrast to the methods presented, artificial neural networks (ANN) are often associated with the ability to develop an intuition for rules and patterns. Ever since the breakthrough in image recognition and speech processing, they have been increasingly used for predicting motions. Previous research has shown that they are clearly superior to conventional approaches regarding certain aspects \cite{Kim2017,Park2018}. Experience-based learning processes make ANN particularly suitable for predicting complex human movement patterns and group dynamics \cite{Alahi2016,Zhang2019}.
Thus, PCMP uses ANN for motion prediction.

Movement predictions exhibit a time reference. In addition, path planning algorithms are defined for graphs. This raises the question of how the obstacles' movements can be mapped into an agent's graph. Two popular discretization algorithms are Probabilistic Road Map (PRM) \cite{vandenBerg2006} and Rapidly Exploring Random Tree (RRT) \cite{LaValle1998} that are often used to discretize a static environment. Previous work has dealt with the question of how these algorithms can be adapted for dynamic environments. In doing so, dynamic obstacles are already taken into account when creating a graph. \cite{vandenBerg2006} is among the first works to investigate the use of deterministic planning algorithms with probability-based representations of the environment. They have added a time dimension to a two-dimensional, spatial PRM. The obstacles' movements are represented as cylinders in this three-dimensional search space and they describe the complete trajectory of an obstacle over time. A graph is created that avoids these trajectories. Then, Dijkstra's algorithm is used to find shortest, collision-free routes. However, a movement's time reference is ignored in their solution. They also assume that the movements are known before the graph is created. In contrast, \cite{Kushleyev2009} presents a novel data structure called Time-Bounded Lattice, i.e. a graph in which the nodes represent different points in time. This graph is adapted to the movement predictions in regular iterations. \cite{Fulgenzi2010} propose a similar approach in which they continuously expand the graph using a RRT. Repetitive discretization processes can be very complex in dynamic environments. Hence, there are alternative data structures such as \emph{Time-Expanded Networks} or \emph{Time-Dependent Graphs} in which repeated discretization processes are not necessary. The originally created graph is only expanded by a time dimension. This means that a graph does not have to be recalculated.

PCMP can be categorized right between motion prediction using ANN, time-dependent graphs and deterministic path planning algorithms. The individual components are usually considered in isolation from one another. We, however, combine the different parts to form a new, predictive and graph-based avoidance solution. The time dimension is the link between the components. It is often neglected in related works, since its relevance only arises in a comprehensive concept such as PCMP. Predictions are probability measurements that pose a collision risk in path planning \cite{Philippsen2006}. This idea is consistently continued in PCMP. The concept of time-dependent risk functions for collision avoidance has received little attention in previous research. There seems to be no work in literature on integrating LSTM motion predictions into time and risk dependent path planning algorithms based on time dependent graphs. In particular by focusing on the time dimension and evaluating collision risks, PCMP contributes to closing this gap.
\section{Predictive Collision Management Path Planning (PCMP)}
\label{sec:pcmp}

In summary, this work's goal is integrating motion predictions into an agent's graph-based planning decisions. On the one hand, the agent should move on the shortest route and on the other hand, collision scenarios should be avoided as far as possible. This compromise describes the central issue of this work. The benefit of motion prediction is only obtained by accepting the risk of collision. Accordingly, safe collision avoidance procedures must be implemented on a local level \cite{Philippsen2006}. PCMP can be seen as part of a multilevel collision avoidance concept as it predicts movements of dynamic obstacles with LSTM models, integrates motion predictions into the routing graph and performs a time- and risk-dependent path planning.

A path's desirability results from its length and the risk of collision. The evasive behavior of the agent can be controlled via a risk parameter that determines the detour's length for which it is worthwhile to avoid a local collision scenario.

While classical shortest paths define problems in a spatial dimension, motion predictions are defined as time-dependent. The time dimension is the framework in which PCMP is implemented.

Terms \textit{collision}, \textit{collision avoidance}, and \textit{collision risk} are integral parts of literature on collision avoidance. We use them with a slightly different meaning. \textit{Collision} does not describe a collision with an obstacle, but a critical situation in which the local collision avoidance procedure must be performed (such a situation is also called \textit{collision scenario} in the following). \textit{Collision risk} is the probability that a path will lead to a critical (\textit{collision}) situation. \textit{Collision avoidance} describes the avoidance of a critical (\textit{collision}) situation.

\subsection{Scenario}\label{sec: scenario}
Agent and obstacles move in a 2D map that is limited on all sides by walls, whereas the agent is confronted with dynamic obstacles on its way. We take the graph, i.e. a discrete representation of space, for granted and we assume it to be undirected with nodes represented by their $(x,y)$ coordinates. In path planning there are different algorithms to discretize a space (PRM, RRT). An important property of these algorithms is that they result in a graph with random structure \cite{Kavraki1996}.

Time-dependent motion predictions are modeled using time-dependent cost functions of the graph's edges. The agent uses a modified A* algorithm for path planning and when executing its plan, it traverses nodes via edges, while obstacles move independently of the graph. The agent can replan its route when reaching a node. If it decides to use an edge, it cannot turn around, in addition, the agent does not have the ability to wait at a node. The agent observes the obstacle's $(x,y)$ positions at regular time intervals. It is assumed that agent and obstacles move at constant speed. In the implementation, a \textit{collision} is assumed to occur when the agent uses an edge that is simultaneously cut by an obstacle.

In literature there is a growing field of research on predicting human movements. In this work, we consider simple motion patterns to limit computational efforts. However, the presented approaches can easily be applied to complex motion patterns. In particular, there are two different patterns. An obstacle type moves linearly towards a randomly chosen target. When the target coordinates are reached, a random target is selected again. A change of direction becomes more likely the closer an obstacle is to a wall. 
Especially at the map's edges uncertainty of future movements increases.
A second type of obstacle oscillates between two points on the map on a parabolic path.

\subsection{Movement Prediction of Dynamic Obstacles with LSTM Models}\label{sec: prediction models}

The prediction model shall predict obstacles' future positions. Basically, a distinction can be made between regression and classification models, whereas the difference lies in the results' structure and information content. Accordingly, integrating predictions into the agent's planning algorithm also differs. While the regression model should be kept as simple and efficient as possible, the more complex classification model should allow a more powerful collision control.

\subsubsection{Basics and Architecture}\label{subsec: architecture}

A model could be trained to predict only the next time step and further predictions build step by step on earlier results \cite{Hahn2018}. PCMP, however, trains a separate model for each time step. This approach causes a higher training effort and is less flexible regarding the prediction sequence's length. However, inaccuracies are not transported over time, which leads to more accurate prediction values at later points in time \cite{Kim2017}. This leads to the question for how many time steps in the future the obstacles' movements should be predicted. The choice of time horizon can be justified by the graph's structure, as the agent can make planning decisions only on nodes. To decide whether an edge will lead to a collision, the prediction horizon must be larger than the usage time of the longest edge. In our scenario, an edge's usage time corresponds to its length. Since in the exemplary used discretization the maximum usage time is $3$ time units, a prediction horizon of $t=4$ was chosen.

\begin{figure}[hpbt]
	\centering
	\includegraphics[width=1.0\linewidth]{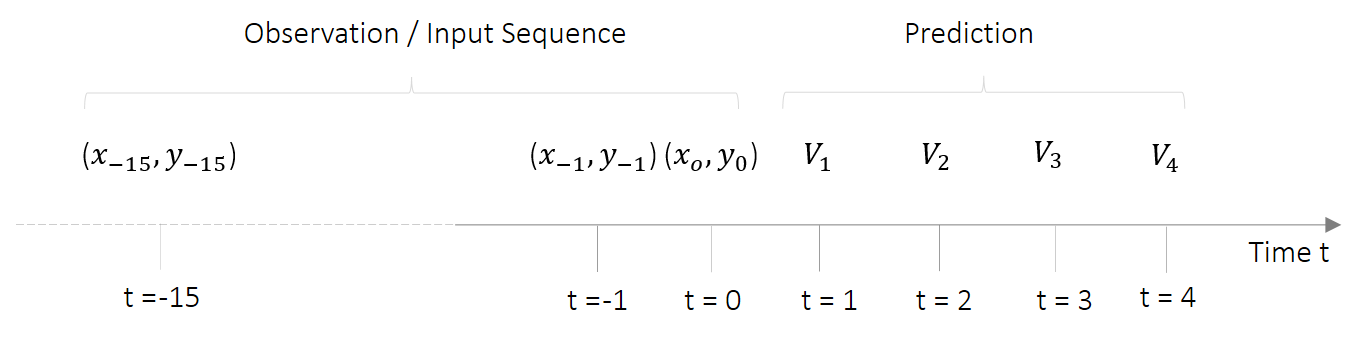}
	\caption [Structure Motion Prediction]{Prediction problem's structure. Based on the last $16$ observations, the next $4$ time steps are predicted. }\label{fig:prediction structure}
\end{figure}

\begin{figure}[hpbt]
	\centering
	\includegraphics[width=1.0\linewidth]{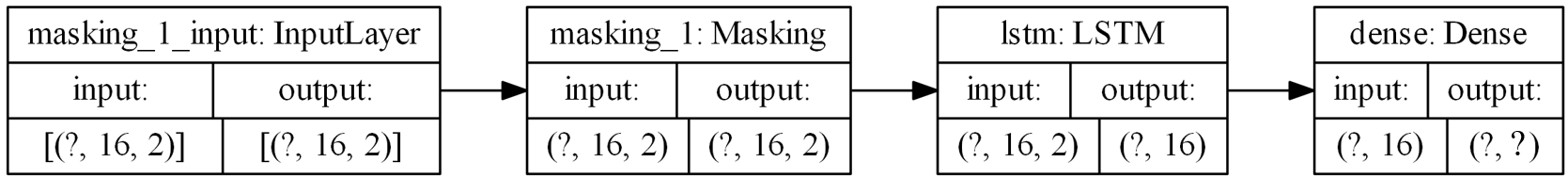}
	\caption [layers prediction model]{Basic structure of the prediction model.}\label{fig:model architecture}
\end{figure}

Fig. \ref{fig:prediction structure} shows the prediction problem's structure. The agent observes the obstacles' positions in its environment at regular intervals, the last observation is at $t = 0$ at position $(x_{0}, y_{0})$. For prediction the last $16$ observations $t=0$ to $t = -15$ are used. The agent should be able to make predictions based on a variable number of observed data. Technically, this property is realized by padding and masking. Training data is generated by simulating and recording the obstacles' trajectories. For each time step $t \in {1,...,4}$ a separate model $V_{t}$ is trained. The training data set consists of $200,000$ sequences of variable length. An input matrix $16\times2$ contains in each line the observed position of the obstacle at corresponding time $t$. Coordinates are normalized to interval $[0,1]$. Fig. \ref{fig:model architecture} shows the model's architecture. Input data first flows through a masking layer so that the padding values are ignored while training. It then passes through a layer of $16$ LSTM cells with ReLU as activation function.

The output layer is formed by a fully connected layer that converts the result into the desired format. Depending on the type of problem (regression/classification), the output layer differs in terms of node number and activation function.

\subsubsection{LSTM Regression Model}\label{subsec: regression model}
\begin{figure}[hpbt]
	\centering
	\includegraphics[width=1.0\linewidth]{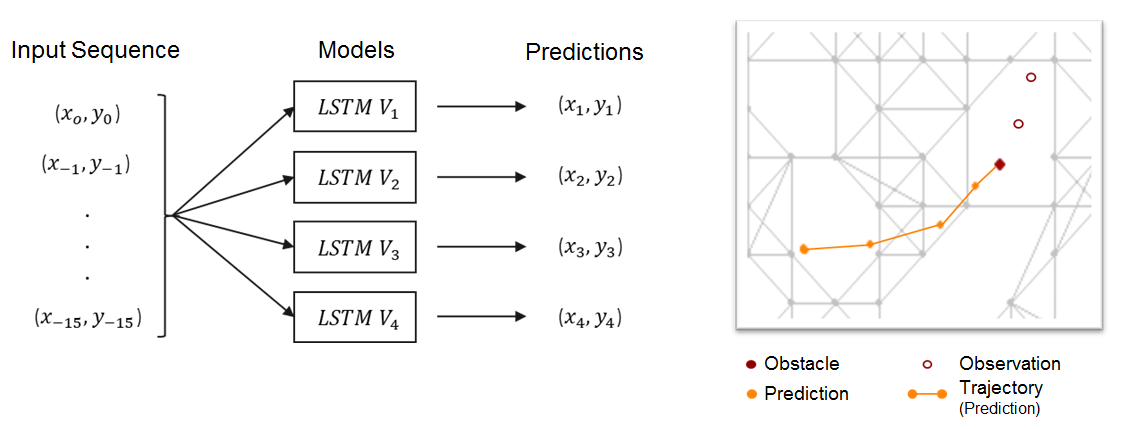}
	\caption [Architecture Regression Model]{Regression model's architecture and movement prediction of an obstacle for $4$ future points. A parabolic movement is predicted.}\label{fig:regression prediction}
\end{figure}

With the regression model, a prediction represents concrete map coordinates where the obstacle is expected to be at corresponding time. Fig. \ref{fig:regression prediction} shows the model's architecture and an example. For each time step $t\in{1,...,4}$ the obstacle's future position $(x_{t}, y_{t})$ is predicted. The ANN's output layer must be adapted to this result format. In concrete terms, two nodes provide a 2D result. With the help of a linear activation function the results can be interpreted as coordinates. The predictions of four separate models are combined to a sequence, interpreted as the obstacle's future trajectory.

\subsubsection{LSTM Classification Model}\label{subsec: classification model}
The classification model's goal is to make probability statements about future movements. A first idea of how to classify movements is provided by literature on occupancy grids \cite{Kim2017} that are typically used to represent the positions of static and dynamic obstacles. The grid cells can be interpreted as categories of a classification problem. By creating a separate occupation grid for each time step, future positions can be represented. 

The core idea of PCMP is the relative definition of movements. An obstacle's future position $(x_{t}, y_{t})$ can be described in relation to the obstacle's current position $(x_{0}, y_{0})$ via $\vec{p_{t}} = (x_{t} - x_{0}, y_{t} - y_{0})$.

First radius $r$ is determined that an obstacle can reach at maximum in the respective time step. This region is equally divided into square cells with the obstacle's current position as center. Individual cells are defined relative to this reference point, thus, one cell describes a set of future positions $\vec{p_{t}}$. The further in the future the prediction is made, the larger is the obstacle's accessible region. The relative structure creates universal categories for the classification problem, which are independent of the obstacle's position and the agent's discretization of space.

\begin{figure}[hpbt]
	\centering
	\includegraphics[width=.7\linewidth]{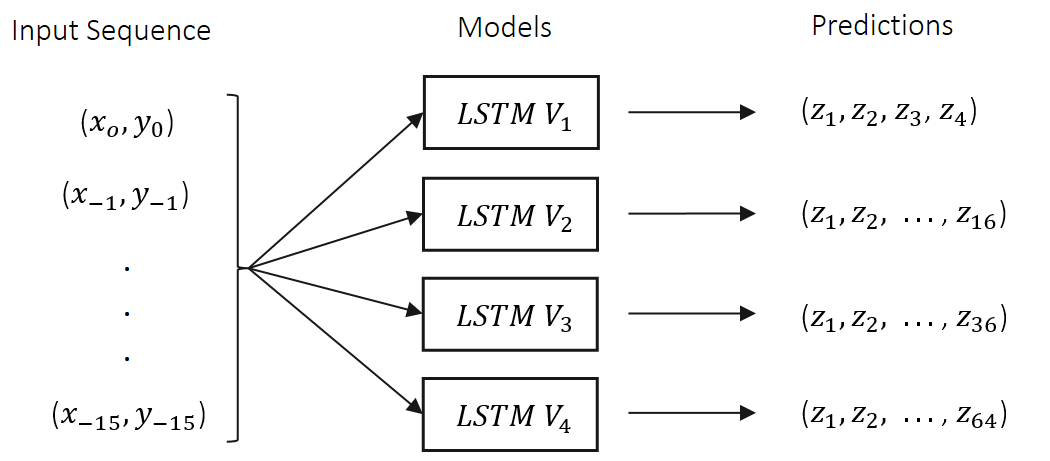}\\\
	\caption [Architecture Classification Model]{Motion prediction via occupancy grids. Different points in time are predicted using different models $V_{t}$. Grid cells $z$ represent the classification problem's categories. 
	}\label{fig:classification problem}
\end{figure}

For each time $t\in{1,...,4}$ a grid with $(2t)^2$ cells is constructed. The grid's granularity can be adjusted according to the motion patterns' complexity and the desired prediction precision. Fig. \ref{fig:classification problem} shows the model's architecture. Result vectors differ in length depending on prediction time $t$. Cells $z$ are categories for an obstacle's future positions. The number of neurons in the output layer of an ANN corresponds to the target vector's length. \textit{Softmax} is used as activation function. Since the result vector's values lie in interval $[0,1]$, they can be interpreted as statements of probability.

\begin{figure}[hpbt]
	\centering
	\includegraphics[width=1.0\linewidth]{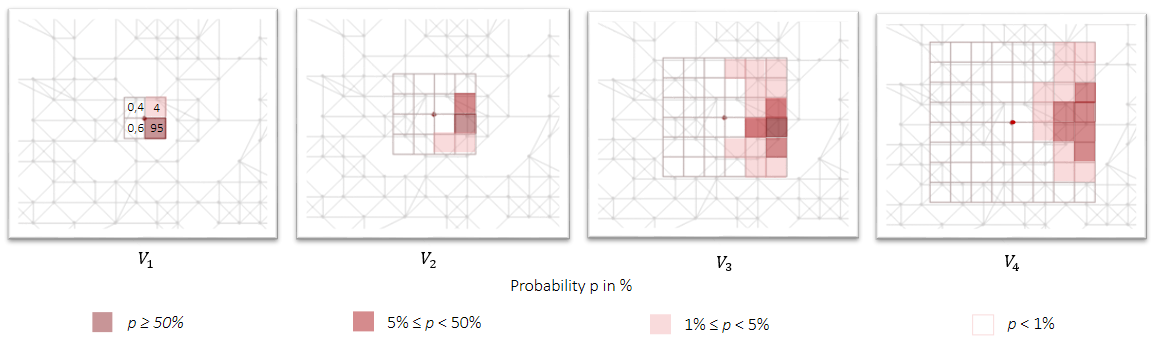}
	\caption [motion prediction with occupancy grids]{Classification results for four points in time $t\in{1,...,4}$. The darker the color, the greater the probability that the obstacle is in this cell at corresponding time.}\label{fig:classification results}
\end{figure}

In contrast to the regression model, no concrete coordinates are predicted, but probabilities of whole cells. In Fig. \ref{fig:classification results} the model $V_{1}$ predicts that at time $t=1$ the obstacle will be located in the dark red quadrant with a probability of $95\%$. A result vector refers to an obstacle's entire accessible region and therefore contains information about all possible trajectories. The most likely trajectory is obtained by connecting the darkest cells. The models can map different trajectories and uncertainties by distributing probabilities. One can see a typical property of the model in the results of $V_{4}$ in Fig. \ref{fig:classification results}. The transition between cells with high and low values tends to be smooth. The predictions are snapshots of the respective points in time.  

\subsection{Integration of Motion Predictions into Time-Dependent Graphs}\label{sec: integration}
The graph is the discretization of an agent's spatial environment and the predictions have a time reference. Although the models are defined for discrete time steps, movements can be interpreted continuously in space and time. A trajectory is the continuous connection between individual predictions and explicitly represents movements. We now create a time-dependent graph that represents motion predictions in a continuous space-time dimension. Using motion predictions, the agent can plan $4$ time steps into future. Within these $4$ time steps, the agent faces a time-dependent planning problem. According to \cite{Foschini2011}, future changes can be modeled with time-dependent edge costs. The result vectors of the regression and classification model differ significantly. Thus, their integration into the graph is solved in different ways. 

\subsubsection{Time Dependent Cost Functions}\label{subsec: cost function}
Using path planning, the agent searches for the optimal route through a graph, which is represented by minimal path costs. Accordingly, the properties of the solution found depend on time-dependent cost functions. In classical shortest path problems, the aim is to minimize distance or travel time between start and destination nodes. In our scenario, the agent should always take the shortest route. If a collision is imminent, the agent should weigh up the risk of collision against alternative solutions and decide according to its risk aversion. The agent uses the path planning algorithm to minimize the sum of the edge weights. The compromise sought is represented at the edges using time-dependent cost functions $c_{v,w}(t) = l + r \cdot k(t)$.

The components of the cost function for an edge $(v,w)$ are the length of an edge $l$ (time independent scalar calculated by the Euclidean distance), a risk parameter for assessing the risk of collision $r$ (to control the agent's risk aversion), and a time-dependent collision function $k(t)$ (representing collision risk of an edge).

The following sections deal with the question of how the prediction models' result vectors can be transformed into time-dependent collision functions $k_{t}$. 

\subsubsection{Graph Integration of Regression Results}\label{subsec: integration_regression}
Through the regression model's prediction, the agent estimates the obstacle's future positions. It can be assumed in a simplified way that an obstacle moves on a direct and linear path to the next predicted position. A trajectory therefore contains both, space and time continuous information. Considering that collisions are defined as intersections with edges while they are used, an intuitive and simple solution approach results. An edge leads to a collision at that point in time by being intersected by the predicted trajectory of an obstacle. As a result, for each time step we get a set of edges that lead to a collision with the obstacle in that time step.

\begin{figure}[hpbt]
	\centering
	\includegraphics[width=1.0\linewidth]{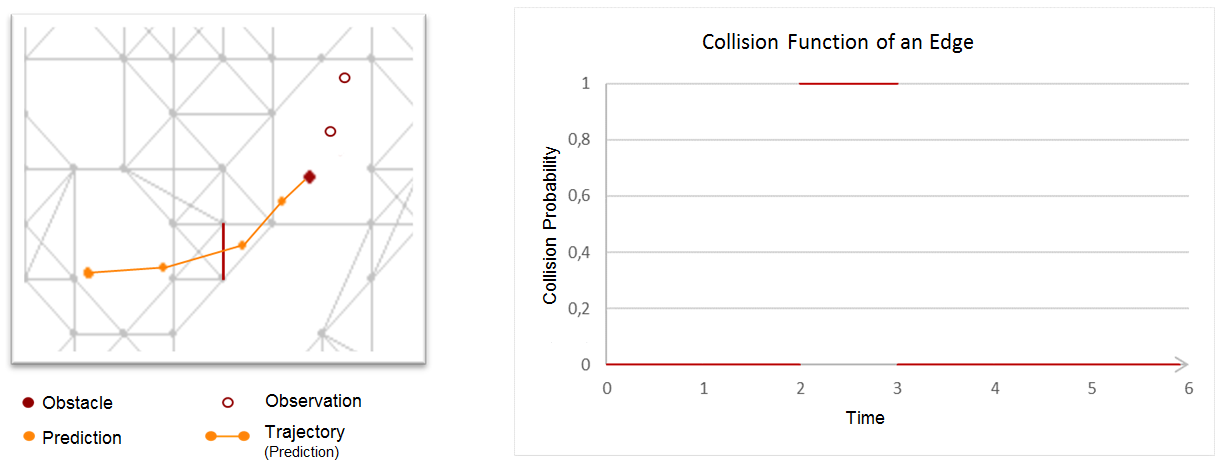}
	\caption [Integration Regression Results and Collision Function]{Integration of regression results into a time-dependent collision function. The red marked edge is cut by the obstacle in time interval $[2,3]$.}\label{fig:regression integration}
\end{figure}

This set is calculated for each obstacle in the agent's environment. It is a binary decision criterion as an edge either leads to a collision or not. Therefore it makes no difference within a time step whether an edge is cut by one or more obstacles. For each time step, the total number of collision edges $K$ is calculated as the union of collision edges of all $k$ obstacles via $K = \bigcup_{i = 1}^k K_{i}$.

Collision times can be integrated in the edges' time-dependent collision functions. They only distinguish between a collision probability of $0\%$ and $100\%$. This results in a time-dependent step function. The step function shown in Fig. \ref{fig:regression integration} leads to a collision between $t=2$ and $t=3$. In contrast to predictions, the created collision functions are continuous in time.

\subsubsection{Graph Integration of Classification Results}\label{subsec: integration_classification}
The classification model provides an occupancy grid for each predicted time step. The translation of cell values in the graphs poses a much more complex problem compared to the regression results. While up to now, exact coordinates of obstacle have been predicted, the accuracy here is limited to whole cells. In contrast to the regression model, no concrete trajectory is predicted. Instead, result vectors contain information about all possible trajectories. It is still possible to derive time-continuous collision functions for individual edges. The problem is first shown for a concrete point in time. In a first step, for each edge, the cells that are traversed by the edge are assigned to it. It is important that several cells can belong to one edge. A calculation rule is used to map assigned cell values to the edge's collision probability at corresponding point in time. Various functions are conceivable at this point. 

\begin{figure}[hpbt]
	\centering
	\includegraphics[width=0.17\textwidth]{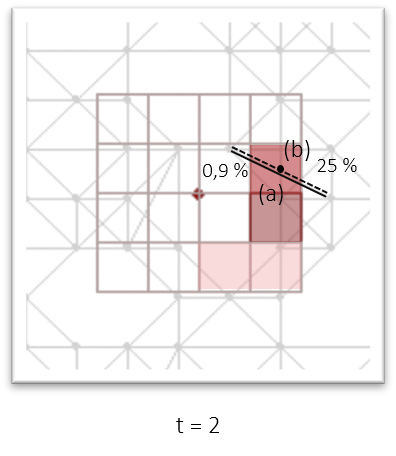}
	\caption [Assignment of cells to edges]{Assignment of different cells to an edge. Solid edge (a) is assigned collision probabilities $0.9\%$ and $25\%$. Dashed edges (b) connect the same end nodes as (a).}\label{fig:edge mapping}
\end{figure}

In contrast to the regression model, all accessible edges are assigned collision probabilities. In order for the agent to judge different attractiveness of paths through the region, edges must be fine-grained distinguishable. An edge's weight could be calculated by selecting the maximum cell probability of the assigned cells. However, this function makes it difficult for the agent to differentiate between edges, since several edges are assigned the same collision probability. An edge with cell values $[25\%, 0.9\%]$ would have the same collision probability as an edge with $[25\%, 25\%]$. The intuition that an edge's high collision value is automatically associated with a conservative avoidance behavior is misleading. The problem could be solved using the average function. However, collision risk of long edges is systematically underestimated. Assume that an agent has the choice between one long \textit{(a)} and two short edges \textit{(b)} (see Fig. \ref{fig:edge mapping}). Both options lead to the same goal and pass through same cells. If the agent uses the two individual edges, the path's collision probability is equal to the two cells' sum ($25.9\%$). If the collision risk of the long edge is equal to the average or maximum cell value, then the collision risk is underestimated ($12.95\%$). The agent would prefer long edges. The exact solution to the problem can be calculated using the proportional distribution. However, it is very complex, which is why following equation is used (note that inaccuracies can also occur). An edge's collision probability at time $k_{t}$ is equal to the sum of cell values $z_{i,t}$ assigned to $n$:

\begin{equation}
k_{t,h} = \sum_{i = 1}^n z_{i,t} \textrm{ for obstacle } h
\label{eq:cell distribution}
\end{equation}

The collision probability in Fig. \ref{fig:edge mapping} is $25.9\%$. The collision probability depends on concrete cell values and the number of cells passed through. The more cells traversed by an edge, the greater its probability of collision. In contrast to the regression problem, this is a statement of probability. It makes a big difference whether an edge may be cut by one or multiple obstacles $n$. The agent first calculates the collision risk for each edge, each time step $t$, and each obstacle $h$ individually $k_{t,h}$. The total collision probability for a time $t$ is calculated using:

\begin{equation}
k_{t,\textit{total}} = 1 - \prod_{i = 1}^n (1-k_{t,i})
\end{equation}

Product $\prod_{i = 1}^n (1-k_{t,i})$ describes the probability that no obstacle will cut the edge. The probability that at least one obstacle collides with the edge follows from the counter probability. In summary, a time-dependent step function for each time step for the collision risk can be derived from this.

\begin{figure}[hpbt]
	\centering
	\includegraphics[width=1.\linewidth]{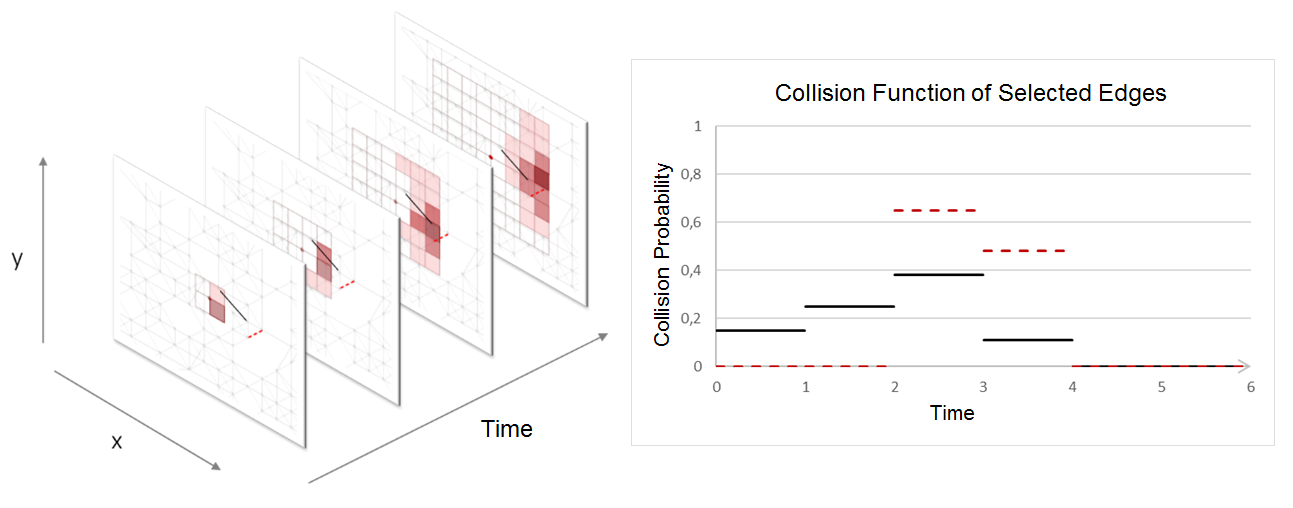}
	\caption [Integration of classification results and collision function]{Integration of classification results in time-dependent collision functions. Red and black function graphs represent time-dependent collision risks for two specific edges.}\label{fig:classification integration}
\end{figure}

Fig. \ref{fig:classification integration} shows the time-continuous risk functions for two examples. The red dotted edge has a collision risk of $0\%$ between time $t=0$ and $t=2$ because it is not reachable within two time steps. In interval $[2,3]$ the edge has a collision risk of $65\%$. For each edge, the collision function for $t>4$ is at $0\%$, as the agent has a horizon into the future of $4$ time steps. At this point it should be noted that the model's accuracy depends on the grid's granularity. Depending on requirements, the procedure can be adapted in terms of accuracy and complexity.

\subsection{Time- and Risk-Dependent Path Planning}\label{sec: algo}
So far, the agent predicts the obstacles' future movements and integrates these predictions into a time-dependent graph. The search for the optimal path in this space-time representation is done by a path planning algorithm, which is introduced in the following.

\subsubsection{Time-dependent A*}\label{subsec: time patterns}
The agent uses a time-dependent A* algorithm based on \cite{Zhao2008} that has been adapted in some key points for the assessment of collision scenarios, with the following pseudocode (start node $s$; target node $d$; planning point in time $t_{0}$; cost function $g(v)$ for path length and collision risk; time period dependent cost function $c_{v,w}(t(v))$ for each edge $(v,w)\in E$; Euclidean distance as heuristic $h(v)$; time function $t(v)$ calculates time of arrival at edge $v$):

\lstset{
	numbers = left,
	mathescape = \true, 
	tabsize = 2,
	basicstyle=\footnotesize,
	xleftmargin=2em
}

\begin{lstlisting}[mathescape] 
$status(s):= marked$, $g(s) = 0$, $t(s) = t_{0}$,
$status(v) := unmarked$ for all $v \in N$ and $v \neq s$.
Let $v$ be a marked node with the smallest $g(v) + h(v)$.
IF $v = z$ GOTO 12
FOR all edges $e(v,w)$ DO:
	IF $status(w) = unmarked$:
		$status(w) = marked, g(w):= g(v)+c_{v,w}(t(v)), p(w):=v$
	ELSE IF $status(w)=marked$ AND $g(w)> g(v)+c_{v,w}(t(v))$:
		$g(w):= g(v)+c_{v,w}(t(v)), p(w):=v$
	END IF
END FOR
$status(v)=finished$, GOTO 2
OUTPUT $g(d)$ and the path $p(s,d)=(d,p(d),p(p(d)),...,s)$
\end{lstlisting}
\vspace*{-5px}
The algorithm's central element are time-dependent edge costs. For this reason, the agent must implement a function $t(v)$ in the planning process that calculates the time point at which an edge is to be used. In \cite{Zhao2008}, path cost to a node $v$ corresponds to the time until the node is reached. In their version, the time of use of an edge is equivalent to the previous path costs. In our work, however, path costs consist of several components. It follows that path costs to an edge $g(v)$ and its time of use $t(v)$ must be calculated with different functions. The next section deals in detail with the calculation of an edge's time of use via time function $t(v)$. This independence is the main difference to \cite{Zhao2008}.

Collision functions considered so far map collision probabilities at different points in time. In the following they will be called time-dependent collision functions. As described in \cite{Demiryurek2010}, an edge's costs depend on the time period in which the edge is used. A time period is defined as an interval over time. Time-dependent collision functions $k_{\textit{time}}(t)$ must be converted to time-period-dependent collision functions $k_{\textit{period}}(t)$.

\vspace*{-5px}
\subsubsection{The Time of Use of an Edge}\label{subsec: time of use}
Basically, the time of arrival at an edge is determined by adding the usage times of all previous edges to a starting value. Assuming that the edge length corresponds to the usage time, as the agent moves with constant speed and that edges have a fixed length, the agent's arrival time at an edge is calculated by summing all $n$ previous edge lengths $l$ and the start time $t_{0}$ via $t(v) = t_{0} + \sum_{i = 1}^n l_{i}$.

The obstacle movement forecasts refer to the observed data and thus to the last observation time $t=0$. The agent only takes planning decisions at nodes, while the time at which it reaches the next node is independent of the obstacles' observation times. The obstacles are detected by sensors at constant time intervals. This means that planning time $t_{0}$ and observation times are different. The upper limit of this discrepancy can correspond to a maximum of one observation interval.

With start value $t_{0}$ the planning time can be synchronized to the forecasts' reference point $t=0$. Value $t_{0}$ is the time span that has passed since the last observation time.

\vspace*{-5px}
\subsubsection{Time Period Dependent Collision Function}\label{subsec: periodic collision function}
The collision functions presented so far represent the risk at different points in time. Considering that an agent's collision risk when using an edge depends on a time period and not on a point in time, it becomes clear that the collision functions set up do not yet have the desired property. What is needed is a cost function which, depending on the time of use $t$, maps the costs over the usage time $d$. If the usage time of an edge is known, the required function can be calculated.

\vspace*{-5px}
\paragraph{Time Period Dependent Collision Function in the Regression Model} 
The sought-after time-period-dependent collision costs can be imagined as a sliding window over the time-dependent collision function. The window size corresponds to the edge's usage time. For the calculation of the collision risk depending on the time period, different mapping rules can be considered. The mapping rule's selection has an influence on the agent's evasion behavior. The agent only knows that the collision will occur at some point in an interval. In this paper, the agent should assume a collision if usage time and collision interval overlap. The time-period-dependent collision risk corresponds to the maximum value of the time-dependent collision function over usage time: $k_{\textit{period}}(t) = \max(k_{\textit{time}}([t:t+d])$.

\vspace*{-5px}
\paragraph{Time Period Dependent Collision Function in the Classification Model}

\begin{figure}[hpbt]
	\centering
	\includegraphics[width=1\linewidth]{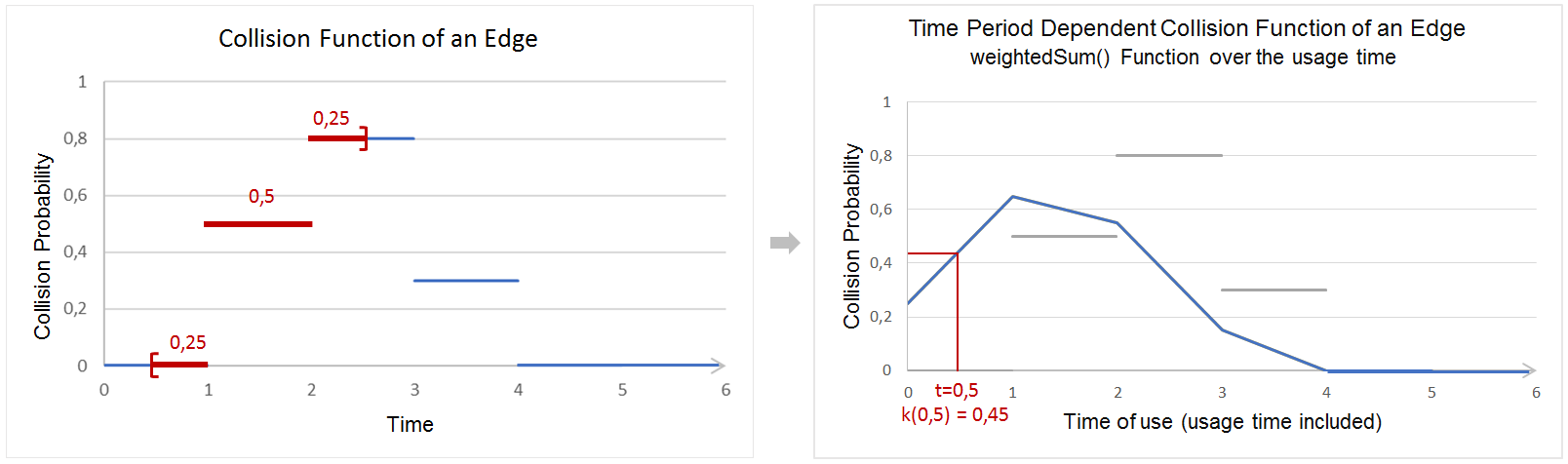}
	\caption [Time Dependent Collision Function - Classification Model]{Translation of the time dependent collision function into a time period dependent collision function for the classification model.}\label{fig:time period classification}
\end{figure}

In contrast to the regression model, classification is based on the obstacle's last observed position and all accessible edges are assigned risk values. The previous developed intuition that the maximum function leads to risk-averse planning decisions does not work in this scenario. The quality of the agent's planning decisions is related to the ability to differentiate between different paths (see Sec. \ref{subsec: integration_classification}). The same argument can be applied to the time dimension. For the agent to be able to judge the attractiveness of an edge between different points in time, as few points in time as possible should have the same level of risk. A step function is not suitable accordingly. The mapping rule here is intended to resolve the staircase structure in the best possible way. The weighted average over the individual risk levels in the usage window fulfills the desired property. On the left-hand side of Fig. \ref{fig:time period classification} the time-dependent collision function with a usage time of $d=2$ is shown. For start time of $t=0.5$ the usage period is marked in red. The risk levels' shares in usage time are also shown.

The weighted average differentiates the collision risk of an edge over the time dimension. The time period dependent collision function is given by $k_{\textit{period}}(t) = \sum_{i=0}^d p(t+i) \cdot k_{\textit{time}}(t+i)$.

Function $p(t)$ calculates proportions of risk levels in the usage time (see Fig. \ref{fig:time period classification}). If an agent uses the displayed edge at time $t = 0.5$ it takes a collision risk of $k_{period}(t) = 0.25 \cdot 0 + 0.5 \cdot 0.5 + 0.25 \cdot 0.8 = 45\%$. At usage time $t=1$ (for a duration $d=2$) the edge has a collision risk of $65\%$. 

\subsection{Risk Parameter $r$}\label{subsec: risk parameter}

PCMP's goal is the intelligent avoidance of local collision scenarios if the detour is worthwhile. The optimal solution is a compromise between risk of collision and shortest route. This compromise depends on the cost of a local collision scenario and the agent's risk aversion, which can be adjusted with risk parameter $r$.

Up to now, only the edges' collision probabilities were considered. Time-dependent A* searches for the path with minimum costs. In order for the algorithm to be able to assess the edges' attractiveness, collision risks must be converted into concrete costs. A simplest cost function is obtained by multiplying collision risk by a risk parameter $r \in \mathbb{R}_{\geq 0}$. The larger the risk parameter, the greater the edge costs. The higher the edge costs, the stronger the agent's incentive to avoid an edge with collision risk. The agent's planning behavior can be adapted to different situations using $r$. Path costs are the sum of the edge costs with $c(t) = l + r \cdot k(t)$. Using risk parameter $r$, the compromise between path length and collision risk can be shifted in both directions as desired. If $r=0$, predictions of dynamic obstacles are ignored.

At this point the development of PCMP is completed. The agent is able to predict obstacles' movements, integrate them into a graph, and make time- and risk-dependent avoidance decisions. Please refer to the \nameref{sec:appendix} for more considerations on the steering effect of $r$ and graphical examples of PCMP's functionality.
\section{Evaluation}\label{sec: eval}
\label{sec:evaluation}

\subsection{Methodology and Reference Values}\label{subsec: methodology}

The agent's avoidance behavior represents a compromise between collision risk and avoidance costs. Accordingly, the number of collisions avoided and the detour's additional length are the key measured values to assess PCMP. To ensure that the experiments reflect the evasive behavior of the specific solution and are independent of the agent's specific planning tasks, a sufficiently large number of path planning operations should be considered. Thus, the experiments consist of $1000$ randomly placed targets for the agent. If the target node is reached, a new target is automatically set.

The path covered consists of the total length of edges used. For each edge we calculate whether a collision has occurred. Both measured values depend on the implemented model and the agent's risk parameter. Besides the risk parameter, the discretization's fineness and number of obstacles play also an important role.

\begin{figure}[hpbt]
	\centering
	\includegraphics[width=0.8\linewidth]{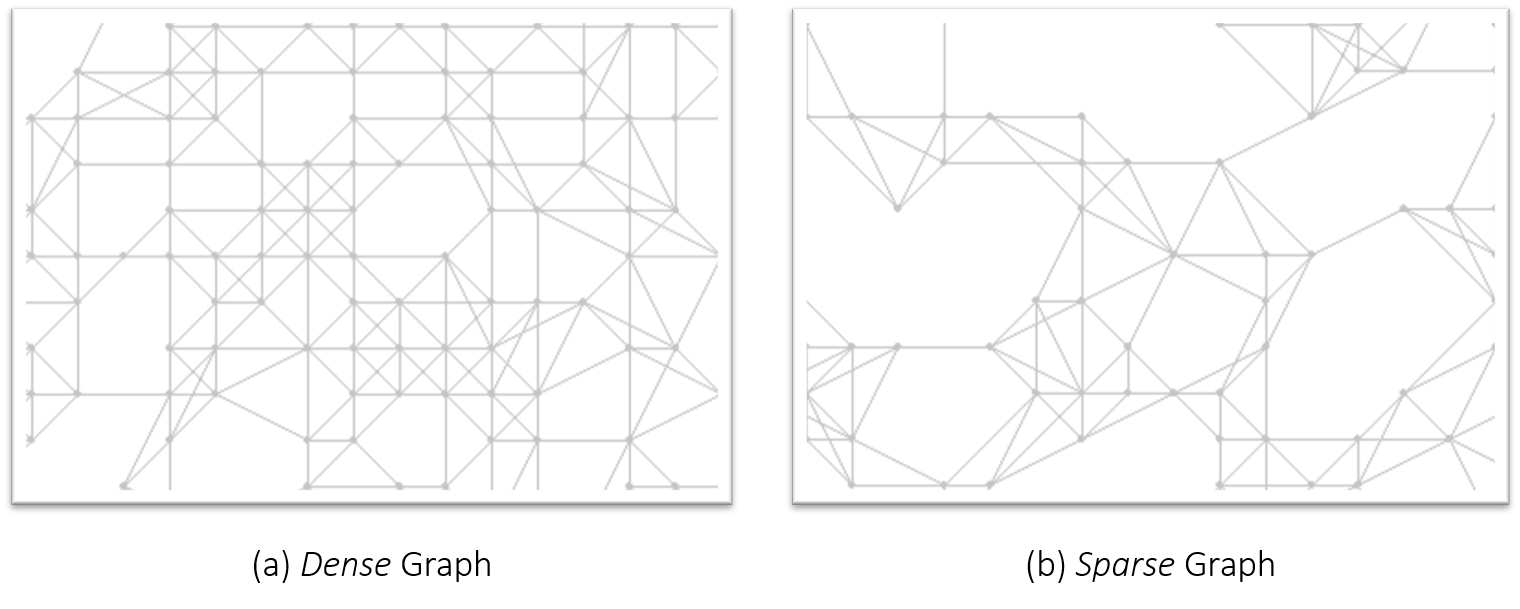}
	\caption [Different Graphs]{Graphs with different accuracy. Graph (a) discretizes the space more finely than graph (b).}\label{fig:sparsedensegraph}
\end{figure}

Fig. \ref{fig:sparsedensegraph} shows snippets of two differently fine graphs. 
Specifically, the algorithm is evaluated in three different scenarios: 

\begin{enumerate}
	\item
	Scenario (a): 8 Obstacles and Discretization/Graph (a) 
	\item
	Scenario (b): 8 Obstacles and Discretization/Graph (b) 
	\item
	Scenario (c): 16 Obstacles and Discretization/Graph (a)
\end{enumerate} 

A simulation over $1000$ paths for an agent without collision avoidance strategy (corresponds to $r=0$) provides reference values which are compared to other settings/scenarios. In scenario (a) having $8$ obstacles, the agent covers a total distance of $\textbf{17117}$ length units. The expected value for an agent without collision avoidance strategy is $\textbf{184}$ collisions. The agent must always reach the same $1000$ targets.

\subsection{Results and Interpretation of the Regression Model}\label{subsec:result regression}

Tab. \ref{regression results} shows the results for PCMP with regression model varying the risk parameter. It is noticeable that the risk parameter and the number of collisions are negatively correlated. The correlation between the risk parameter and the detour's length is positive. 

\begin{table}[htp]
	\centering
	\caption [results regression model]{Collision control via regression as a function of risk parameter.}\label{regression results}
	\includegraphics[width=.99\linewidth]{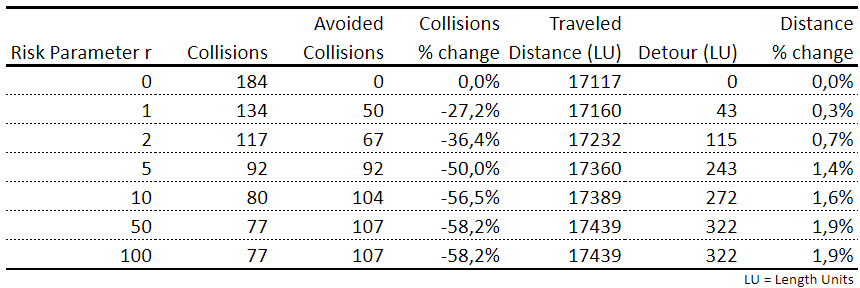}
\end{table}

\begin{figure}[hpbt]
	\centering
	\includegraphics[width=1\linewidth]{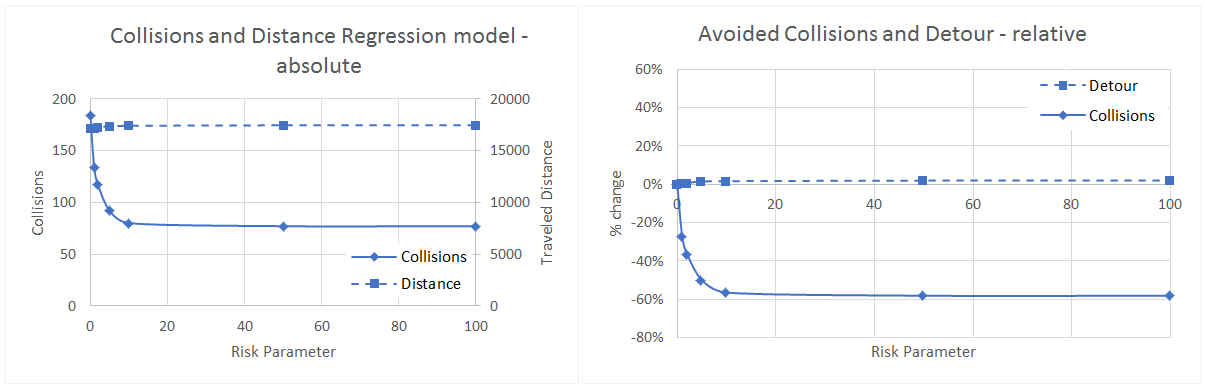}
	\caption [Results regression model]{Collision control via regression depending on risk parameter. Measurements are shown in absolute values (left) and as relative changes to reference value $r=0$ (right).}\label{fig:regressioneval}
\end{figure}

The table shows that with a risk parameter of $r=1$, the agent avoids $50$ of $184$ collisions. In the given graph, the agent can avoid a total of $27.2\%$ of collisions by a detour of $0.3\%$. From a value of $r=50$ on, no additional control effect is apparent in the agent's avoidance behavior. In the given scenario, the agent can avoid almost $60\%$ of collisions. It achieves this reduction with a total detour of about $2\%$. Fig. \ref{fig:regressioneval} shows the number of collisions and the distance traveled in absolute numbers on the left-hand side. On the right-hand side, the percentage deviations of the measured values from the reference values are shown.

The results can be explained by the graph's structure and the implemented integration approach (see Sec. \ref{subsec: integration_regression}). The regression model punishes affected edges very selectively. Due to the nodes' high degree of interconnection, there are often many alternative routes to a destination node, which limits the influence of the risk parameter. Furthermore, the agent has a strong dependence on the prediction model's quality which might not always be correct.

\subsection{Results and Interpretation of the Classification Model}\label{subsec:results classification}

Tab. \ref{classification resultseval} presents the results of the classification model for different risk preferences of the agent. Due to the risk parameter's macro effect (described in the \nameref{sec:appendix}) it is to be expected that the collision avoidance behavior can be controlled in a more differentiated way via the parameter value.   

\begin{table}[htp]
	\centering
	\caption [results classification model]{Collision control via classification depending on various risk parameters.}\label{classification resultseval}
	\includegraphics[width=.99\linewidth]{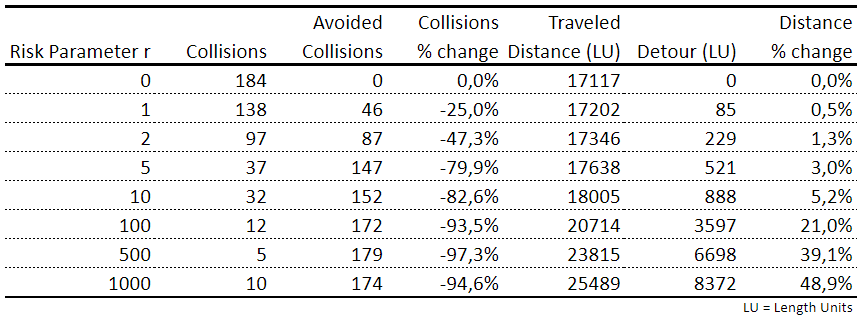}
\end{table}

\begin{figure}[hpbt]
	\centering
	\includegraphics[width=1\linewidth]{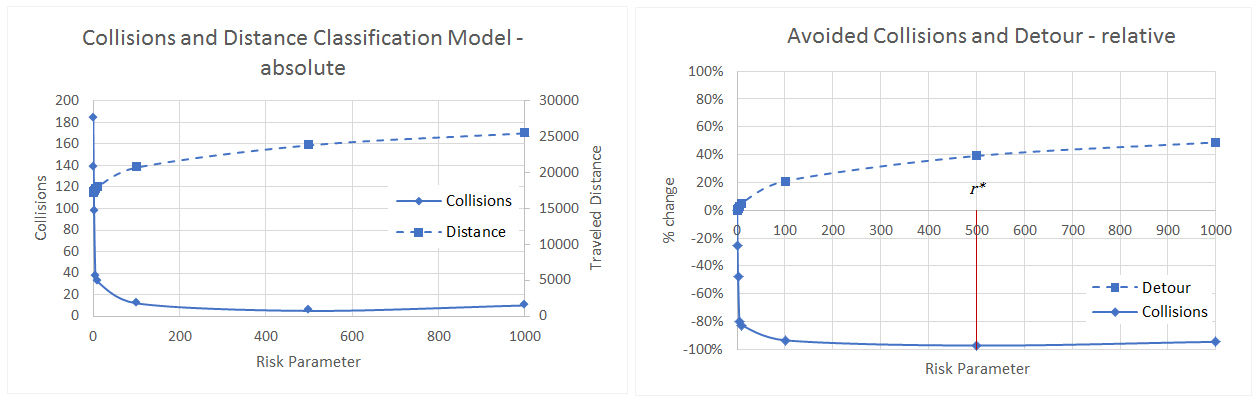}
	\caption [Results classification model]{Collision control via classification depending on various risk parameters. Absolute values (left) and relative changes to reference value $r=0$ (right).}\label{fig:klassifkationeval}
\end{figure}

Basically, following correlation can be observed in the results for $r\in [0.500]$: The greater the agent's risk aversion, the more collisions are avoided and the longer the detours are. A risk-affine agent ($r=1$) will only deviate from the shortest route if the probability of a collision is very high and the alternative solution means only a minimal detour. This agent avoids about $25\%$ of collisions with a detour of $0.4\%$. It assesses a collision risk of $100\%$ in the same way as a risk-averse agent ($r=100$) assesses a collision risk of $1\%$. The risk-averse agent avoids about $93\%$ of the collision scenarios and accepts a detour of $21\%$. Fig. \ref{fig:klassifkationeval} shows the simulation results of the classification solution.

Using the risk parameter, the number of collision scenarios can be reduced to a very small level. The larger the risk parameter, the greater the arc around an obstacle. Here, the agent also depends on the prediction's certainty as it determines the region's size to which costs are added. Results show that not $100\%$ of collisions can be avoided which can be explained by the limited planning horizon.

Results also show that the number of collisions for $r=1000$ increases slightly compared to $r=500$. Because of the macro effect, the agent avoids obstacles increasingly (see \nameref{sec:appendix}). On these long detours, the agent encounters obstacles that it would not have encountered if it had not avoided the first obstacles. The collision graph (blue line) shown in Fig. \ref{fig:klassifkationeval} has a minimum at $r^*$. In interval $[0,r^*]$ the agent finds the best possible combinations of detour and collision risk. In summary, the agent can avoid up to $95\%$ of collision scenarios via the classification model.

\subsection{Alternative scenarios}\label{subsec:result validation}

\paragraph{Scenario (b) - Sparse Graph:}

The discretization of space influences the agent's ability to avoid a collision. In a coarser discretization, the agent has fewer alternative solutions in its planning decisions. It is therefore to be expected that detours tend to be longer. Tab. \ref{regressionresultsparse} and Tab. \ref{classificationresultsevalsparse} show the results for scenario (b).

\begin{table}[htp]
	\centering
	\caption [Results Regression Model - Sparse Graph]{Regression PCMP for Scenario (b): Sparse Graph.}\label{regressionresultsparse}
	\includegraphics[width=.99\linewidth]{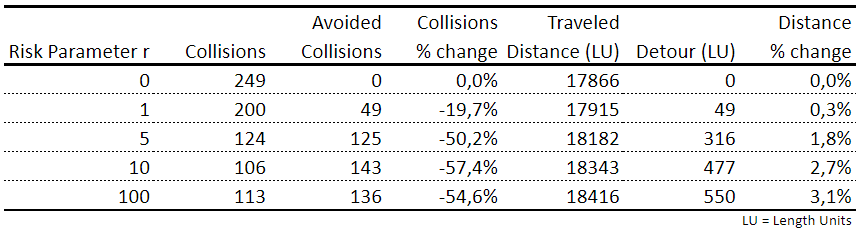}	
\end{table}

\begin{table}[htp]
	\centering
	\caption [Results Classification Model - Sparse Graph]{Classification PCMP for Scenario (b): Sparse Graph.}\label{classificationresultsevalsparse}
	\includegraphics[width=.99\linewidth]{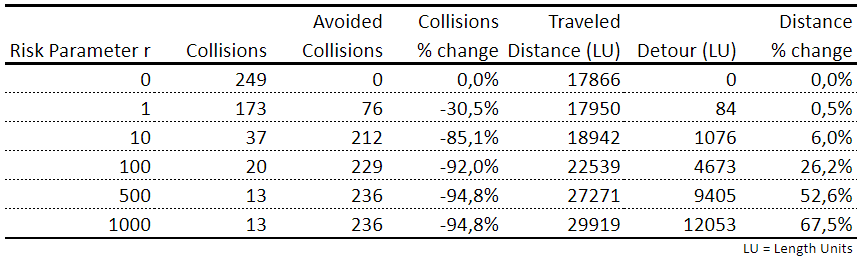}
\end{table}

Results show that PCMP works for a coarse discretization as well as for a fine one. Thus, PCMP's functionality seems independent of a concrete graph.
However, the graph's structure influences the relationship between avoided collision scenarios and detour. Results in Tab. \ref{classificationresultsevalsparse} show a significant increase in detour compared to scenario (a) with the same percentage of collisions.

\begin{figure}[hpbt]
	\centering
	\includegraphics[width=0.5\linewidth]{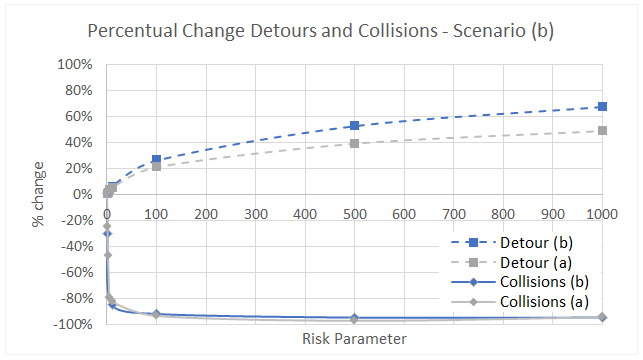}
	\caption [Results Classification Model - Sparse Graph]{Comparison of Classification Model between sparse (b) and dense discretization (a).}\label{fig:EvaluationSzenariob}
\end{figure}

The effect of different discretizations is shown in Fig. \ref{fig:EvaluationSzenariob} for the classification model. It shows that the percentage detour for $r=1000$ is $18.6\%$ longer in scenario (b) than in scenario (a) (see Tab. \ref{classification resultseval}: $48.9\%$ detour). This avoids approximately the same number of collisions ((a): $94.6\%$ and (b): $94.8\%$). This difference can be attributed to the graph's fineness. With a fine graph the agent has many possibilities to avoid a collision risk with a very small detour. In scenario (b), the agent has fewer possibilities to move around in space. Therefore, alternative routes tend to lead to longer detours.

\paragraph{Scenario (c) - Many obstacles}

In this scenario, the agent is faced with a large number of obstacles. Specifically, there are $16$ obstacles moving (compared to $8$ obstacles before). Tab. \ref{regressionresults16} and Tab. \ref{classification resultseval16} show the results.

\begin{table}[htp]
	\centering
	\caption [Results Regression Model - Many Obstacles]{Regression PCMP for Scenario (c): Many Obstacles.}\label{regressionresults16}
	\includegraphics[width=.99\linewidth]{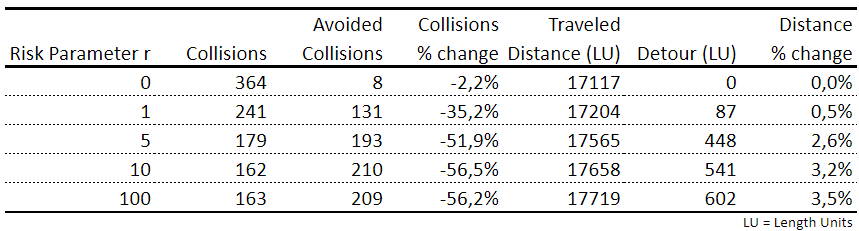}
\end{table}

\begin{table}[htp]
	\centering
	\caption [Results Classification Model - Many Obstacles]{Classification PCMP Scenario (c): Many Obstacles.}\label{classification resultseval16}
	\includegraphics[width=.99\linewidth]{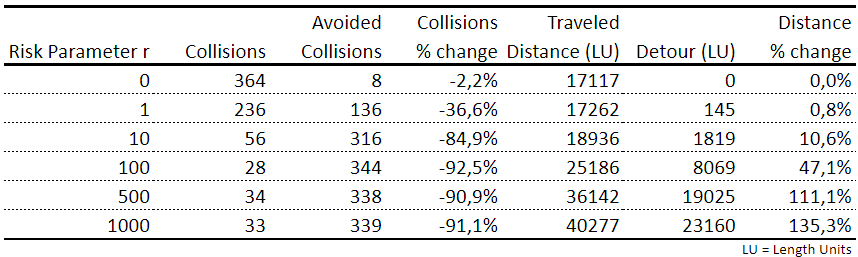}
\end{table}

\begin{figure}[hpbt]
	\centering
	\includegraphics[width=0.5\linewidth]{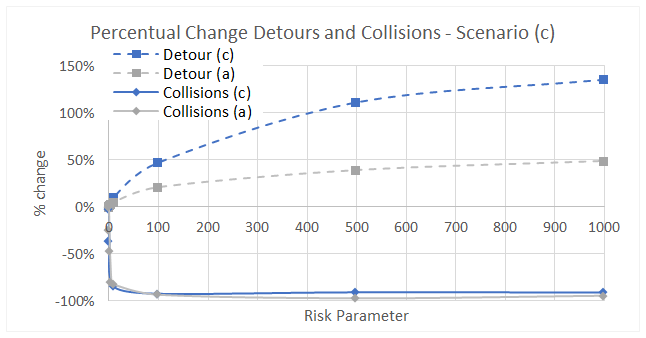}
	\caption [Results Classification Model - Many Obstacles]{Comparison of results of classification model with 8 (a) and 16 (c) obstacles.}\label{fig:EvaluationSzenarioc}
\end{figure}

It becomes clear that PCMP is also usable in environments with many obstacles. The agent can successfully navigate between obstacles and make risk-based avoidance decisions. Regression results are consistent with the other scenarios and underline the effectiveness of this approach. Fig. \ref{fig:EvaluationSzenarioc} shows the effect of the increase in obstacles on the measured values. It is easy to see that the agent evades a similar percentage of obstacles as in scenario (a). 

The sharp percentage increase in detour is due to the larger number of obstacles. Logically, the agent has to avoid more collisions and therefore has to take larger detours. If the agent moves in an environment with many obstacles, it is in danger of getting stuck in an endless loop. This can occur if the agent is trapped between trajectories of several obstacles that oscillate in a parabolic shape. If the agent detects such a dead end, it increases its risk preference to resolve the situation locally. This way, a path's risk of collision gets less weight and the agent gets closer to its target.
\section{Conclusion}
\label{sec:conclusion}

We have presented a collision control concept for predictive path planning called PCMP.
The time dimension is the link between movement predictions and time-dependent path planning decisions. In this context, the role of time receives litte attention in literature. Predictions are probability statements that pose a collision risk in path planning. The consideration of time-dependent risk functions has often been neglected in previous work. From this gap follows the necessity, relevance and contribution of our work.
By using LSTM models, the agent learns an intuition for the movement patterns of dynamic obstacles. Movement predictions are formulated as trajectories (regression model) and occupancy grids (classification model). The models differ in terms of information content and complexity. The central concept for integrating motion forecasts into graphs is time-dependent edge costs. A modified A* algorithm enables searching optimal paths in the space-time dimension. The solution of the path planning algorithm represents a compromise between collision risk and path length (which can be controlled via a risk parameter). On the one hand, the agent should move along the shortest path, and on the other hand, it should avoid collision scenarios as best as possible. The results show that PCMP successfully integrates this decision into the agent's planning process. PCMP is therefore suitable as a predictive collision management algorithm for time and risk-dependent path planning.

\subsection{Limits and Outlook}
\label{subsec:limits}

In order to keep the regression approach as simple as possible, uncertainties were not taken into account in the predictions. The avoidance behavior can be controlled much more differently if edges are subject to collision risks. Dealing with uncertainty is a central research direction in literature on ANN \cite{Khosravi2011}. Some of the methods presented are suitable for converting regression results into probability-based statements \cite{Qiu2019}. As a result, uncertainty in movement predictions can be explicitly taken into account in the planning process. Such an expansion would on the one hand increase the regression model's control options and on the other hand reduce the dependence on the LSTM model's quality.

We assumed that the agent moves in every time step. In many situations, however, waiting is an interesting alternative to avoid collisions. The search algorithm would have to be expanded to include the additional alternative. \cite{Silver2005a} presented first solutions to that. The extension decouples travel time and path length. This leads to further interesting questions when searching for solutions. In our work, a collision was defined as the intersection of obstacles and the agent's path at the same time interval. On the one hand, the definition of a collision as a distance to the obstacle is more realistic. On the other hand, the agent can better differentiate between the costs over time. Differentiating the collision risk within a time interval would enable more precise collision control.

We considered planning behavior for a relatively small graph. For larger graphs, traditional planning algorithms reach their limits due to large search spaces. We mentioned a variety of modified versions of Dijkstra's algorithm to increase the algorithms' efficiency. These optimizations can also be applied to the planning algorithm of our work. By separating the time-dependent and static search, both parts can be optimized separately. The D* Light algorithm is suitable for static searches, since the graph does not change outside the time horizon and repeated searches can be significantly accelerated.

Based on the basic ideas of this work, various design decisions and assumptions allow a large scope for further research. The generic structure of PCMP offers space to investigate alternative solutions for individual steps. One example is the architecture of motion prediction. A separate model was trained for each time step. In a future work, the time dimension could also be considered as a parameter in a global LSTM. The introduction of different speeds, more complex LSTM models, and finer assignment grids represent possibilities for extension.

\bibliographystyle{ACM-Reference-Format}
\bibliography{main}

\clearpage
\appendix
\section{Appendix}\label{sec:appendix}
In the appendix the steering effect of the risk parameter $ r $ shall be further outlined as well a graphical intuition for the functionality of PCMP shall be developed.
Specifically, the avoidance behavior of an agent for a concrete planning decisions is considered.

\paragraph{Interpretation of the Risk Parameter:}
The steering effect and interpretation of the risk parameter is divided into a micro and a macro effect.
The micro-effect describes the effect of $ r $ in relation to the risk assessment of an edge.
The macro effect results from the fact that in the classification model whole regions are assigned collision probabilities.  

In the regression approach, the risk parameter $r$ can be interpreted as follows: Assuming that the best path to date certainly leads to a collision and there is an alternative route with additional costs of $k < r$, the agent decides to take the detour.
There is no risk control in the true sense of the word, since the agent only distinguishes binary between the collision edges.
The risk parameter expresses the value of an avoided collision for the agent.
The larger the parameter, the longer the detours that the agent takes to avoid a collision.

In the classification model, the edges are assigned collision probabilities.
Assume that there are exactly two possible routes to the target point.
The agent is on the most attractive route and predicts a collision risk of $ 1\% $.
The alternative route causes $ 8 $ additional cost units.
The agent has a parameter value of $r=1000$.
The collision risk of $ 1\% $ increases the cost of the original route by $ 10 $.
The agent will choose the detour and avoid the risk.
It shows risk-averse behavior.
A riskier agent has a parameter value of $r=100$.
In this case, the collision risk of the current path must be at least $ 8\% $ for the agent to decide to take the detour.
If one abstracts from individual edges to a macro perspective, a further control effect of $ r $ is obtained in the classification model.
In contrast to the regression model, risk values are assigned not only to individual edges but to all accessible edges of the agent.
The predictions divide the environment into cells with high and low risk.
Typically, the transition between these regions is fluid (see Figure \ref{fig:classification results}).
This characteristic in the occupancy grid becomes accessible for the agent via the risk parameter.
The higher the risk parameter, the higher the edge costs.
From the property described in the predictions, it follows that the agent takes a greater arc around regions with a high risk of collision.
The higher the risk parameter, the greater the safety distance between the planned path and the future movements of the obstacle.

\begin{figure}[hpbt]
	\centering
	\includegraphics[width=.7\linewidth]{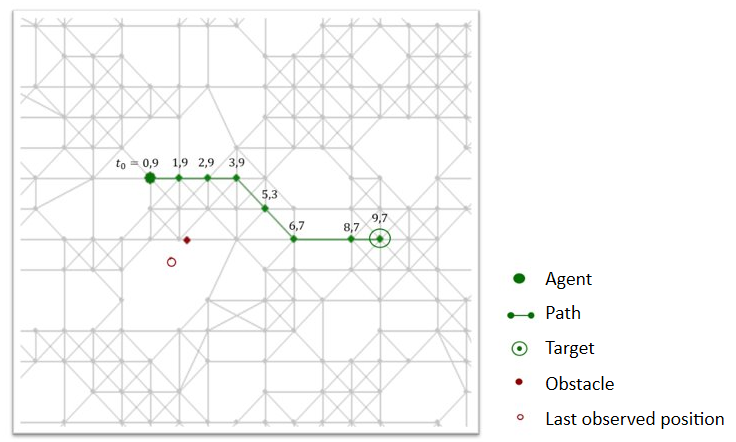}\\\
	\caption [alternative situation example]{Example of a concrete planning decision.}\label{fig:collision behaviour example}
\end{figure}
A concrete planning problem is shown in Figure \ref{fig:collision behaviour example}.
It shows an agent that is in the vicinity of an obstacle.
The agent must consider in its planning decision that the obstacle (red dot) has moved on since the last observation (red circle).
Since the last observation $0.9$ time steps have passed ($t_{0} = 0.9$).
The times of use of the edges are entered at the nodes.
The time of use of an edge is decisive for the calculation of collision costs and the avoidance behavior.
The next sections show the different models in action.  

\begin{figure}[hpbt]
	\centering
	\includegraphics[width=1.0\linewidth]{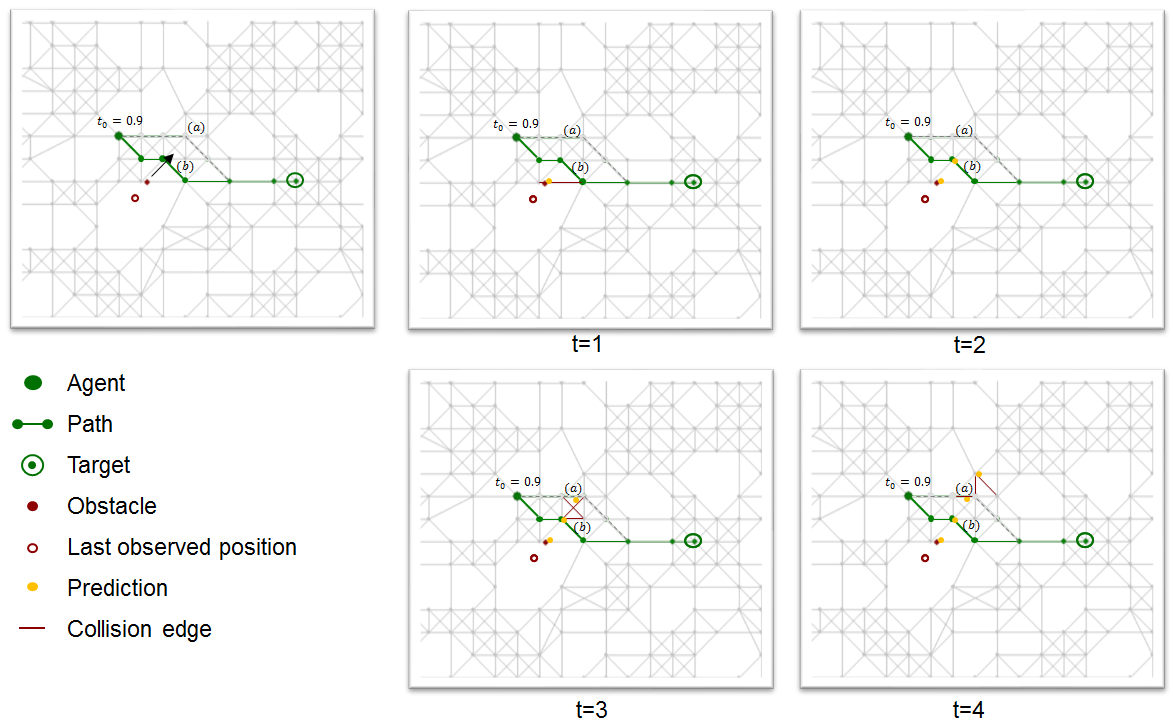}\\\
	\caption [evasion behavior regression model]{ Evasion behavior of an agent with regression model and $r=100$. Snapshots of the edge weights at different points in time.}\label{fig:collision behaviorregression model}
\end{figure}

\paragraph{Evasion Behavior Regression Model:} Figure \ref{fig:collision behaviorregression model} shows the evasion behavior of the agent for the regression model with a risk parameter of $r=100$.

At first glance, it can be seen that the agent barely overlaps/runs behind the obstacle.
Apparently, the agent predicts a collision on its previous path (including edge (a)) and avoids it.
It is noticeable that the new solution has the same length as the old one.
The agent can avoid a collision scenario without taking a detour.
To be able to understand the agent's evasive behavior, different snapshots of the graph are shown at different points in time.
The future positions of the obstacles are predicted as coordinates.
The edges that lead to a collision are marked in red in the corresponding time step.
They cause additional costs of $r=100$.
Edge (a) leads to a collision in time interval $[3,4]$ and edge (b) between $[1,2]$.
Both paths intersect the predicted trajectory of the obstacle.
The difference is the time of use of the edge.
The agent calculates the time of arrival at an edge in the planning process.
Edge (a) is by the agent in the interval $[2.9,3.9]$.
The agent predicts that the edge in this interval leads to a collision and avoids it.
In contrast, in the new solution the agent uses edge (b) between $[2.3,3.7]$.
Using the cost function of the edge, the agent knows that the obstacle has already passed the edge at least $ 0.3 $ time units ago. 
In the example shown, the central properties of the regression model can be seen.
The agent only occupies the edges with collision costs that are cut by the obstacle.
The agent does not distinguish between edges that are near the predicted trajectory and edges that are far away from it.
The typical avoidance behavior of the agent follows from this binary decision criterion.
The detours to avoid a collision are as minimal as possible.

\begin{figure}[hpbt]
	\centering
	\includegraphics[width=1.0\linewidth]{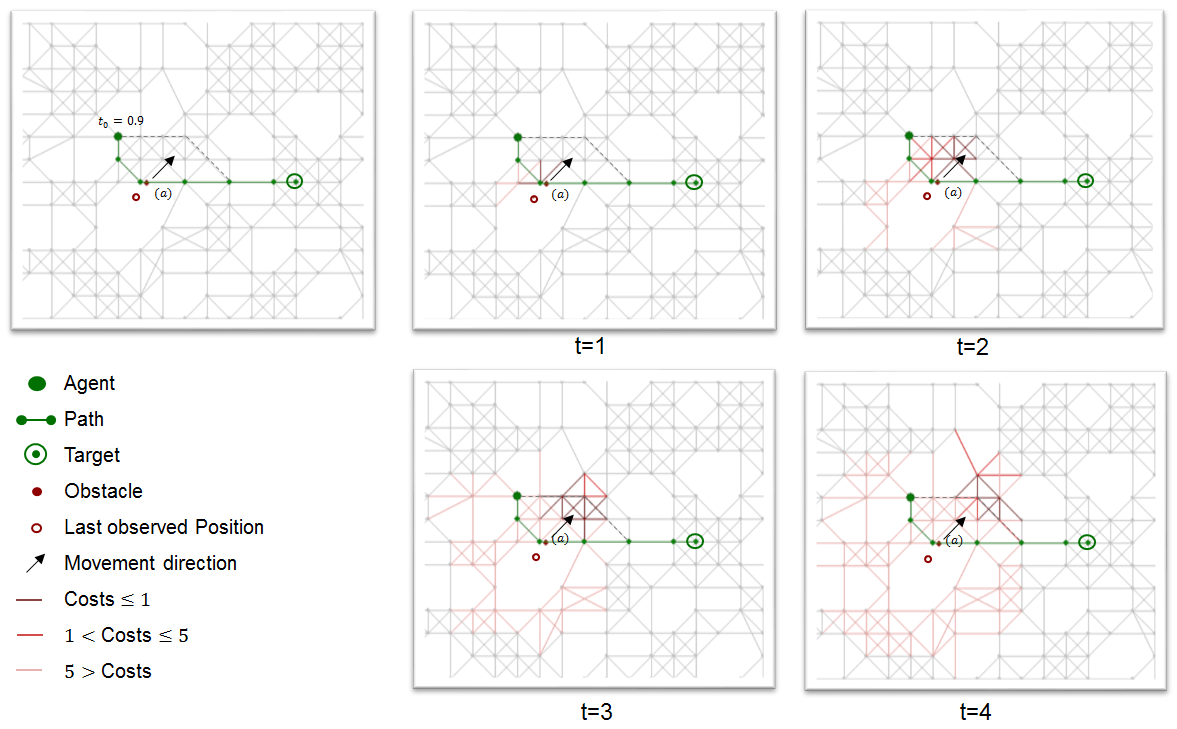}\\
	\caption [Risk-Affine Evasive Behavior Classification Model]{ Evasion behavior of an agent with classification model and $r=100$. Snapshots of the edge weights at different points in time.}\label{fig:collision behaviorClassification model100}
\end{figure}

\paragraph{Evasion Behavior Classification Model:} With the classification model, the agent can not only avoid individual edges, but also control the length of the detour. Therefore, the avoidance behavior is considered for a risk-affine ($r=100$) and a risk-averse ($r=1000$) agent. Figure \ref{fig:collision behaviorClassification model100} shows the result for $r=100$. 

In the classification model, all accessible edges are assigned a collision risk.
Accordingly, all edges cause additional collision costs.
The different costs of the edges are indicated by color shades.
The darker the edge, the higher the costs.
The future course of movement of the obstacle is easy to see.
The region of risk (dark edges) moves over the graph with time.
The edge (a) has a high risk of collision in the time step $t=1$ and a low risk in $t=4$.
Decisive for the agent is the cost of an edge at the time of use.
Similar to the regression model, the agent narrowly overlaps/run behind the obstacle.
The new path is slightly longer than the old one.
The agent avoids the dark edges. The new path consists only of light edges.
These edges have a small risk of collision.
It is not worthwhile for the agent to take an even greater detour to avoid this residual risk.
The example shows the balance between collision risk and path length. \\

\begin{figure}[hpbt]
	\centering
	\includegraphics[width=1.0\linewidth]{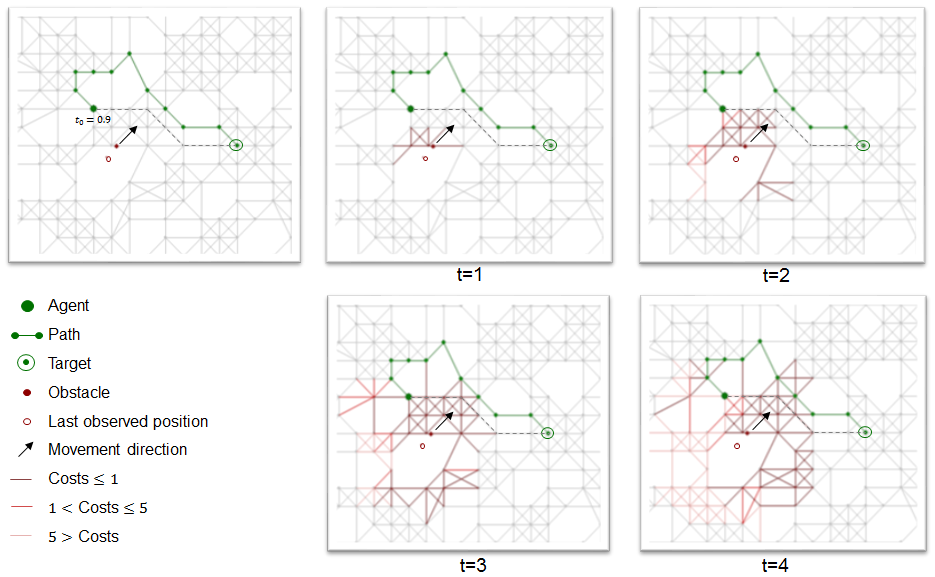}\\\
	\caption [risk-averse evasion behaviour classification model]{ Evasion behavior of an agent with classification model and $r=1000$. Snapshots of the edge weights at different points in time.}\label{fig:collision behaviorClassification model1000}
\end{figure}

A risk-averse agent ($r=1000$) has a much smaller tolerance limit regarding the collision risk of an edge. Figure \ref{fig:collision behaviorClassification model1000} shows the result for $r=1000$. 

At first glance, it can be seen that a large proportion of the accessible edges cause high costs.
The predicted collision probabilities are the same as in the previous example.
However, the risk assessment has changed.
An edge with a collision probability of $ 1\% $ and $r=1000$ causes the same costs as a risk of $ 10\%  $and $r=100$.
Accordingly, the risk area (dark edges) is much larger than before.
The extent of the risk area (dark edges) can be controlled by the risk parameter.
As a consequence, the agent avoids the obstacle in a large arc.
This observation was previously introduced as a macro effect.
The larger the risk parameter, the more risk-averse the agent is and the more widely it avoids an obstacle.  \\

\begin{figure}[h]
	\centering
	\includegraphics[width=1.0\linewidth]{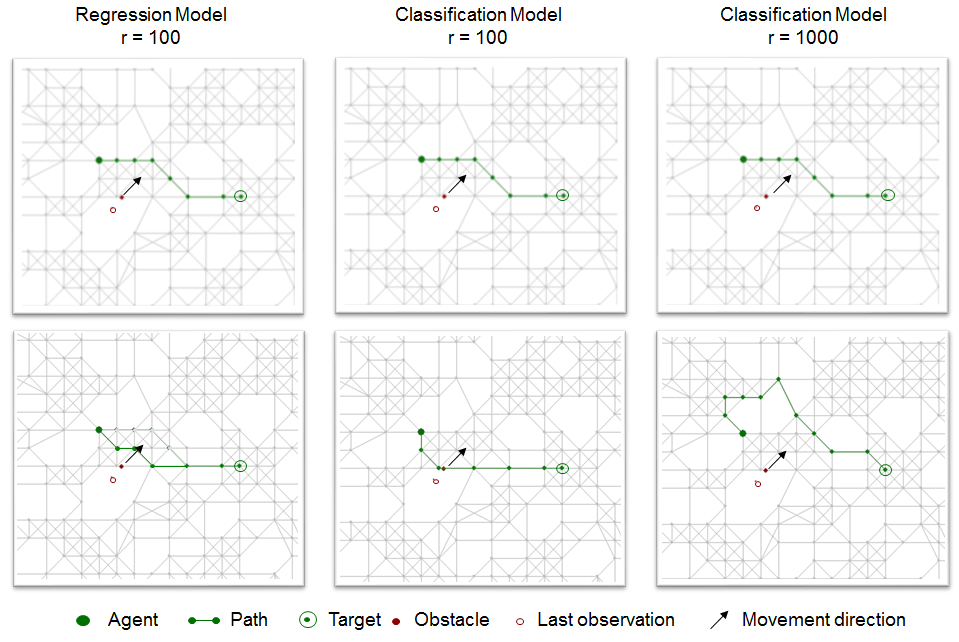}\\
	\caption [comparison of avoidance behavior]{ Results of the PCMP in different variants for a concrete planning decision.}\label{fig:comparison of avoidance behavior}
\end{figure}

\paragraph{Comparison:} Figure \ref{fig:comparison of avoidance behavior} shows the different results side by side.
The different models cause different collision behavior.
In the regression model, the agent evades an obstacle as closely as possible.
In the classification approach, the distance between path and obstacle is controllable.
The concrete evasion decision of the agent always represents a balance between the risk of collision and the shortest route.
With the risk parameter, the agent can prioritize one of the two factors. 

\typeout{get arXiv to do 4 passes: Label(s) may have changed. Rerun}
\end{document}